\pdfoutput=1

\documentclass[11pt]{article}

\usepackage[final]{acl}

\usepackage{times}
\usepackage{latexsym}

\usepackage[T1]{fontenc}

\usepackage[utf8]{inputenc}

\usepackage{microtype}

\usepackage{inconsolata}

\usepackage{graphicx}

\usepackage{algorithm}
\usepackage{algpseudocode}
\usepackage{adjustbox}
\usepackage{booktabs}
\usepackage{multirow}
\usepackage{soul} %
\usepackage{listings}
\usepackage[most]{tcolorbox}
\usepackage[capitalize]{cleveref}
\usepackage{makecell}
\usepackage{authblk}  %

\newcommand*{\yellowmarker}{\includegraphics[scale=0.09]{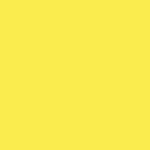}}%

\newcommand{\myyellow}[1]{\hl{#1}}
\newcommand{\ourmethod}{LAQuer}

\lstdefinestyle{promptStyle}
{
    basicstyle={\footnotesize\ttfamily},%
    numbers=none,
    numberstyle=\footnotesize,
    xrightmargin=1.5em,
    showstringspaces=false,
      showspaces=false,
        showtabs=false,
    tabsize=2,
    breaklines=false,
        flexiblecolumns=true,
        escapeinside={<@}{@>},
          breakatwhitespace=true
}

\newtcblisting{mylisting}[1]{
  enhanced,
  listing only,
  boxrule=0.8pt,
  sharp corners=downhill,
  top=0mm,
  bottom=0mm,
  left=2mm,
  right=0mm,
  boxsep=0mm,
  colframe=black,
  colback=white,
  listing options={
    style=#1
  }
}

\definecolor{instructionsColor}{rgb}{0.1, 0.5, 0.1}

\usepackage[notes=true, done=false, later=true, camera]{dtrt}
\newcommand{\eh}[1]{\dtcolornote[Eran]{blue}{#1}}

\title{\ourmethod{}: Localized Attribution Queries in Content-grounded Generation}

\author{
    \textbf{Eran Hirsch$^{1}$} \qquad
    \textbf{Aviv Slobodkin$^{1}$} \qquad
    \textbf{David Wan$^{2}$} \\[0.7em]
    \textbf{Elias Stengel-Eskin$^{2}$} \qquad
    \textbf{Mohit Bansal$^{2}$} \qquad
    \textbf{Ido Dagan$^{1}$}
}
\affil{$^{1}$Bar-Ilan University \qquad{} $^{2}$UNC Chapel Hill}
\affil{\tt \{hirsch.eran, lovodkin93\}@gmail.com \\ \tt \{davidwan, esteng, mbansal\}@cs.unc.edu \quad{} dagan@cs.biu.ac.il}

\begin{document}
\maketitle
\begin{abstract}

Grounded text generation models often produce content that deviates from their source material, requiring user verification to ensure accuracy. Existing attribution methods associate entire sentences with source documents, which can be overwhelming for users seeking to fact-check specific claims. In contrast, existing sub-sentence attribution methods may be more precise but fail to align with users' interests. In light of these limitations, we introduce \textbf{L}ocalized \textbf{A}ttribution \textbf{Quer}ies (\ourmethod{}), a new task that localizes selected spans of generated output to their corresponding source spans, allowing fine-grained and user-directed attribution. We compare two approaches for the \ourmethod{} task, including prompting large language models (LLMs) and leveraging LLM internal representations. We then explore a modeling framework that extends existing attributed text generation methods to \ourmethod{}. We evaluate this framework across two grounded text generation tasks: Multi-document Summarization (MDS) and Long-form Question Answering (LFQA). Our findings show that \ourmethod{} methods significantly reduce the length of the attributed text. Our contributions include: (1) proposing the \ourmethod{} task to enhance attribution usability, (2) suggesting a modeling framework and benchmarking multiple baselines, and (3) proposing a new evaluation setting to promote future research on localized attribution in content-grounded generation.\footnote{\url{https://github.com/eranhirs/LAQuer} \eh{add code}}
\end{abstract}

\textit{``ChatGPT can make mistakes. Check important information.'' --- ChatGPT interface}

\section{Introduction}\label{sec:Introduction}
\eh{to see if this motivation is included: sentence-level citaitons are problematic because sentences might have a mix of faithful and unfaithful information, resulting in the model not providing citaions at all (cite Buchmann) }
Grounded text generation aims to produce content based on specific sources, whether retrieved—such as in retrieval-augmented generation (RAG) \citep{RAG, ram-etal-2023-context}—or user-provided.
Yet, model outputs frequently diverge from these sources, resulting in factual inaccuracies, or `hallucinations' \citep{Mishra2024FinegrainedHD}.
To address this, users often need to manually review retrieved documents to ensure the accuracy of generated claims. 
This in turn has driven a growing interest in \textit{attributed} text generation \citep{Thoppilan2022LaMDALM, Menick2022TeachingLM, bohnet2023attributed}, which incorporates supporting evidence or citations into the output, thereby enhancing model reliability and helping mitigate potential factuality errors.

\begin{figure}[t]
    \centering
    \includegraphics[width=\linewidth]{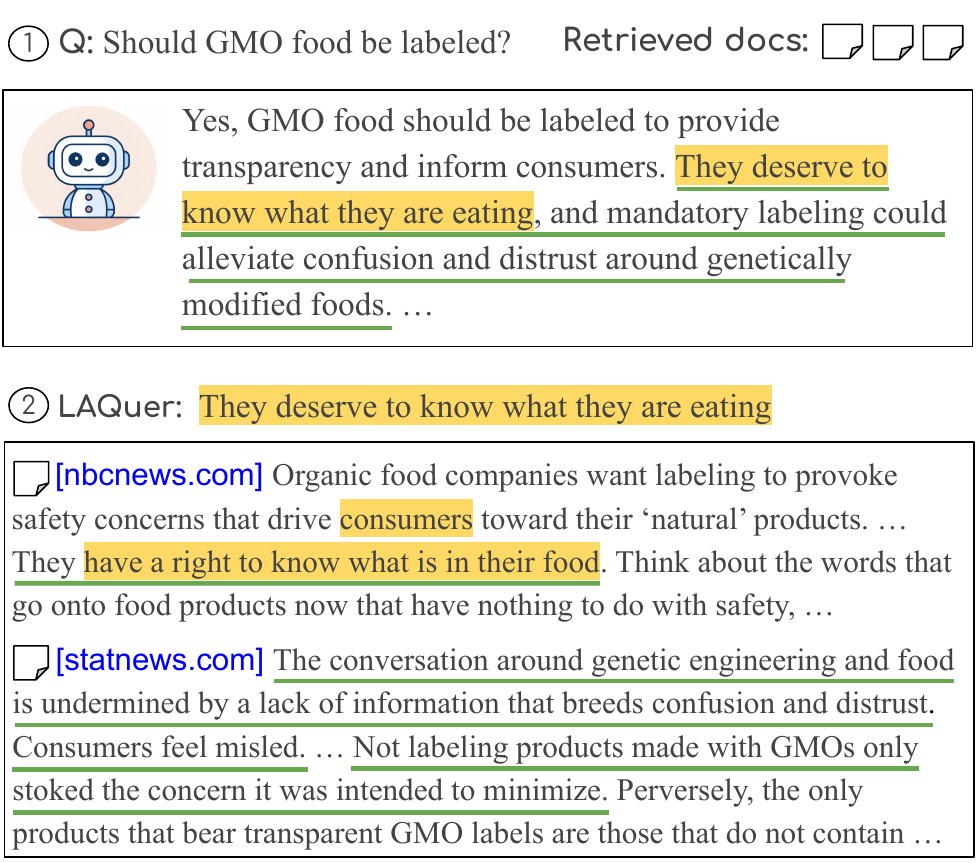}
    \caption{\textbf{Top}: example RAG scenario. \textbf{Bottom}: our Localized Attribution Queries (\ourmethod{}), where the attribution is constructed per user query, highlighted in yellow. Existing sentence-level attribution methods, underlined in green, can often be disorienting and lengthy.}
    \label{fig:iola_example}
\end{figure}

While attributed text generation enhances transparency by providing citations, its effectiveness depends on how easily users can interpret these attributions, as shown in \cref{fig:iola_example}.
Most existing attribution methods associate each generated \textit{sentence} with its corresponding attributions \citep{gao-etal-2023-enabling, slobodkin-etal-2024-attribute}. For example, the output sentence underlined green is attributed to many spans in the source document, also underlined green. Yet, in practice, users often seek to fact-check specific details rather than an entire sentence (e.g., the highlighted fact in \cref{fig:iola_example}). 
As sentences typically contain multiple facts \citep{min-etal-2023-factscore}, sentence-level attribution requires readers to examine both the full sentence and its sources before assessing factual accuracy of a single fact. 
For instance, in \cref{fig:iola_example}, the highlighted fact is attributed by the first source, while another within the same sentence is linked to the second source. As a result, users must review the entire sentence and all cited sources to verify this single fact.

In this work, we introduce a more precise attributed generation task, which we call \textbf{L}ocalized \textbf{A}ttribution \textbf{Quer}ies (\ourmethod{}), 
that links specific spans in generated text to their corresponding source spans.
Each query consists of pre-selected output spans, or `highlights' (e.g., the highlighted span in the top part of \cref{fig:iola_example}), while the response identifies the relevant source spans (e.g., the highlighted spans in the bottom part of \cref{fig:iola_example}). 
Since queries can vary from single words to full sentences, this approach generalizes existing attribution methods while enabling targeted attribution.

\begin{figure*}[t]
    \centering
    \includegraphics[width=\linewidth]{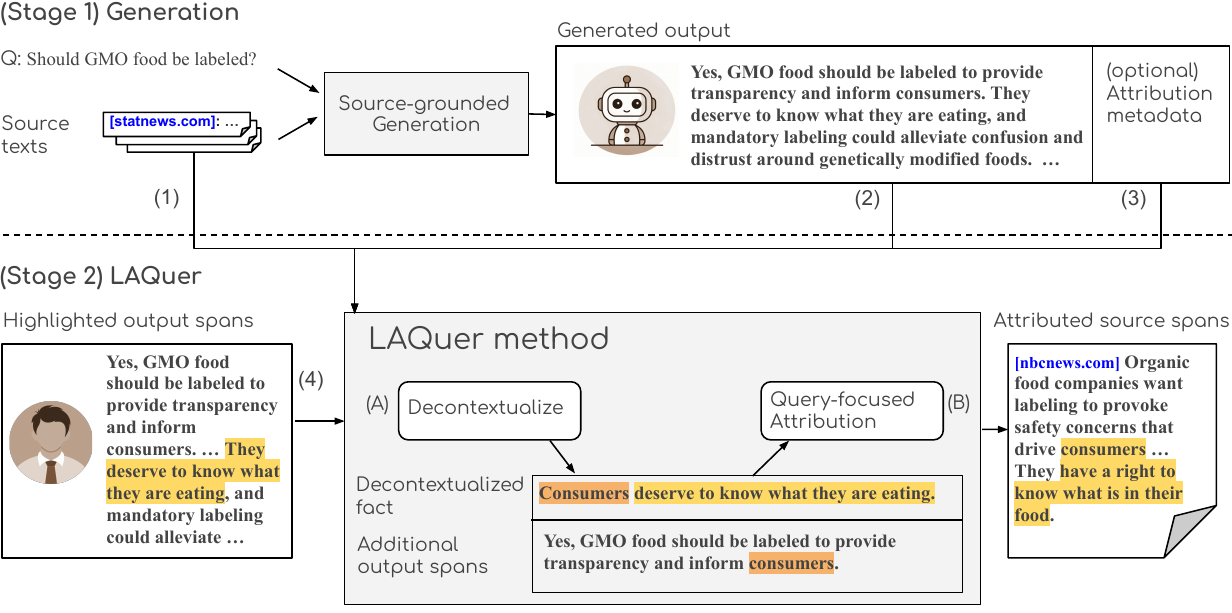}
    \caption{Overview of our \ourmethod{} framework. The top section illustrates the generation of an output based on identified source texts, either provided as input or retrieved. The bottom section represents the \ourmethod{} task, where output spans are attributed back to their source texts, enabling users to verify the provenance of individual pieces of information. The inputs to our proposed \ourmethod{} approach are labeled (1) to (4). In Step (A), the highlighted spans are transformed into a decontextualized fact along with its corresponding output spans. In Step (B), the user's query is attributed to relevant source texts, enabling precise fact verification. %
    }
    \label{fig:contributions}
\end{figure*}

We model the \ourmethod{} setting as a framework consisting of two processing stages, illustrated in \cref{fig:contributions}. %
First, a source-grounded generation system produces text expected to be supported by identified source texts. Some generation methods may include attribution metadata, mapping output segments to supporting source spans. For example, in \cref{fig:iola_example}, a sentence-level attribution method can attribute the second sentence to the texts underlined green.
In our experiments (\cref{sec:benchmark}), we benchmark \ourmethod{} using three generation approaches: one without attribution and two contemporary attributed-generation methods. In the second stage, users request localized attribution by highlighting spans that correspond to a fact of interest. The \ourmethod{} task then identifies the exact supporting source spans for the given highlight. This second stage is composed of two steps: (A) decontextualization of the user's query, and (B) query-focused attribution. The decontextualization step converts the highlighted fact to a stand-alone decontextualized statement, for which source attribution can be more easily sought in an unambiguous matter. For example, ``\emph{They}'' in \cref{fig:iola_example} refers to ``\emph{consumers}''. In such cases, attributions should account for the decontextualized meaning, e.g., that ``\emph{They}'' is correctly attributed to ``\emph{consumers}.''
The query-focused attribution step searches for the supporting source spans for the decontextualized statement. For the query-focused attribution, we compare two approaches: one that prompts a large language model (LLM) to produce the alignment and another that uses the internal representations of the model to align phrases \citep{phukan-etal-2024-peering}. If attribution metadata from the generation step is available, it is leveraged to narrow the search space. For example, instead of scanning the entire source document in the figure, our approach can focus on the spans underlined green.

For evaluation, to simulate user interaction in this process, our methodology involves decomposing the generated output into atomic facts using LLMs \citep{min-etal-2023-factscore}, which are subsequently aligned with output spans. The \ourmethod{} task can then be applied to any type of generation, unlike previous work which focuses on datasets annotated with sub-sentence alignments \citep{phukan-etal-2024-peering, qi-etal-2024-model, cohen-wang2024contextcite}.
Our experimental setup includes two grounded generation tasks, Multi-document Summarization (MDS) and Long-form Question Answering (LFQA). A key finding is that \ourmethod{} methods can significantly reduce the length of the attributed text. Overall, \ourmethod{} remains a challenging task, particularly in attributing decontextualized facts.
In total, our contribution in this work is enumerated as follows:

\begin{enumerate}
    \item We propose Localized Attribution Queries (\ourmethod{}) as a task to improve the accessibility of attributions for users.
    \item We introduce a novel modeling framework for the \ourmethod{} setting and benchmark various baselines. We demonstrate their potential to enable targeted attribution while maintaining accuracy. %
    \item We establish a new evaluation setting that encourages future research on localized attribution in content-grounded generation.
\end{enumerate}

\section{Background}\label{sec:Background}

Hallucinations produced by LLMs have attracted increasing interest in generating attributed text. The task of \textit{attributed text generation} requires models to generate summaries or answers that cite specific evidence for their claims \citep{gao-etal-2023-enabling, Thoppilan2022LaMDALM, Menick2022TeachingLM, bohnet2023attributed}. When considering the granularity of the attribution, there are two key factors: the granularity of the summary or answer (i.e., the output) and the granularity of the source text (i.e., the input).
The standard level of output granularity is sentence-level \citep{gao-etal-2023-enabling, slobodkin-etal-2024-attribute}. Some work focuses on sub-sentence attribution, based on the internal representations of a model \citep{phukan-etal-2024-peering, qi-etal-2024-model, ding2024attentiondependencyparsingaugmentation} or manipulation to the input \citep{cohen-wang2024contextcite}. Similarly, input granularity can vary between pointing to the entire response \citep{Thoppilan2022LaMDALM}, documents \citep{gao-etal-2023-enabling}, snippets \citep{Menick2022TeachingLM}, paragraphs or sentences \citep{buchmann-etal-2024-attribute}, and spans \citep{schuster-etal-2024-semqa,phukan-etal-2024-peering,qi-etal-2024-model,ding2024attentiondependencyparsingaugmentation,cohen-wang2024contextcite}.

The above methods provide fixed pre-determined attributions, that often do not correspond most effectively to the specific scope of output information for which attribution is sought. Some systems provide attributions for longer output spans, requiring the user to examine irrelevant source segments \citep{gao-etal-2023-enabling,slobodkin-etal-2024-attribute}, while others provide only partial attributions for narrow output spans, requiring the user to look around the attributed source spans for complete supporting information \citep{phukan-etal-2024-peering, qi-etal-2024-model, ding2024attentiondependencyparsingaugmentation}.
Our work is the first to explore user-initiated attribution queries across variable scales, introducing a novel evaluation methodology to assess their effectiveness.

\section{Localized Attribution Queries}\label{sec:IOLA_task_definition}

The \ourmethod{} task assumes as input a generation $o$ grounded in source documents $D$. For instance, in \cref{fig:iola_example}, the answer to the question \emph{``Should GMO food be labeled?''} is generated based on two source documents.
A key aspect of this task is the inclusion of `highlights', which are specific parts of the generated output that are marked by the user. These highlights indicate a fact that the user wants to verify or examine within the source. The user conveys the fact of interest by selecting the spans in the output that best express it. For example, in the figure the highlighted span is: \emph{``They deserve to know what they are eating.''} Importantly, the user may not care about other claims made in the same sentence, such as \emph{``labeling could alleviate confusion and distrust.''} Formally, we are given a set of highlighted output spans $o_1,\ldots,o_n$ where each span may range in length from a single word to the entire generated output.
The goal of the \ourmethod{} task is to provide the highlighted source spans $s_1, \ldots{}, s_m$ that support the fact expressed in these highlights. 

Within this setting, we aim that our \ourmethod{} task definition would capture the following desiderata: 

\paragraph{1)$\:$ User-initiated Attribution Queries.}
Most attribution methods provide `fixed', pre-determined attributions, meaning that attribution is generated alongside the output, only allowing users to explore it afterward \citep{gao-etal-2023-enabling, slobodkin-etal-2024-attribute, phukan-etal-2024-peering}.
However, we point out that users are often interested in checking the attribution only for a limited subset of facts within the generated output, and it is not possible to predict a user's specific interests in advance. 
\ourmethod{} requires developing methods that can dynamically provide attribution for any arbitrary fact of interest, which the user highlights in the output.

\paragraph{2)$\:$ Source and Output Localization.}
\citet{slobodkin-etal-2024-attribute} introduce the Locally Attributable Text Generation task, where the goal is to provide the user with concise \textit{source} spans necessary to verify a \textit{complete} output sentence; in other words, the goal is to provide \emph{localized}, precise input spans.
In this work, we also consider the localization for the other side of the attribution, which is the \textit{output} localization. Instead of complete output sentences, we work with output spans. Formally, the concatenation of the source spans should contain only the necessary content to support the information conveyed by the output spans.

\paragraph{3)$\:$ Output Decontextualization.} Given a contextualized claim $c$ extracted from some text $r$, \citet{choi-etal-2021-decontextualization} define a decontextualized claim $m$ as one that uniquely specify entities, events, and other context such that the claim $c$ is now interpretable. 
In our setting, it is likely that the highlights provided by the user are contextualized. 
For example, the output spans in \cref{fig:iola_example} mention \emph{``They,''}
which refers to the consumers mentioned in the previous sentence. 
However, the user did not highlight \emph{``consumers,''} because it is redundant and can be inferred from \emph{``They.''}
Accordingly, source spans should correspond to a decontextualized version of the output. 
For example, in \cref{fig:iola_example}, the source from \texttt{nbcnews.com} must explicitly include \emph{``consumers''} to avoid ambiguity. 
Only including \emph{``they''} in the source spans would be problematic, as it lacks a clear referent and could lead to misinterpretation or false attributions.
Formally, we denote the decontextualized meaning of the output spans in the context of the complete output as $I(o_1,\ldots{},o_n|o)$. The source spans should express the decontextualized meaning of the output spans, $concat(s_1, \ldots{}, s_m)\models{I(o_1,\ldots{},o_n|o)}$.

\section{\ourmethod{} Modeling Framework}\label{sec:frameworks}

The \ourmethod{} setting, as defined above, inherently involves two processing stages, illustrated in \cref{fig:contributions}. In the first stage, a source-grounded generation system generates a user-requested text, such as a summary or a long-form answer to a question, based on provided documents. 
This system may also include attribution metadata, mapping output segments to supporting source segments. For example, in \cref{fig:iola_example}, the generation system could output the sentence-level attribution underlined green.
In our experiments (\cref{sec:benchmark}), we evaluate \ourmethod{} using three generation methods: one without attributions and two recent attributed-generation approaches.

In the second stage, users who read the generated text can request localized attribution for specific facts by highlighting relevant spans. The \ourmethod{} task then identifies the exact supporting source spans for the highlighted facts. %
Specifically, during stage 2, the \ourmethod{} input consists of the following: (1) the source documents, based on which the output text was generated; (2) the generated output text; (3) the attribution metadata (if available); (4) the output spans highlighted by the user, which are assumed to correspond to a particular fact in the output text, for which attribution is sought.

Given these inputs, our proposed \ourmethod{} method first performs a decontextualization step (A), which converts the input highlights into a coherent standalone sentence. Next, in the attribution step (B), we search for the supporting source spans that provide evidence for the decontextualized statement, where we explore two alternative methods for this step (prompt- and internals-based). This step leverages the attribution metadata from the generation step, if available, while also incorporating the extended highlights. 
These two steps are described in detail below.

\begin{table*}
    \small
    \centering
    \setlength{\tabcolsep}{0.5em}
    \renewcommand{\arraystretch}{1.2}
    \adjustbox{max width=\textwidth}{
    \begin{tabular}{p{10cm}p{5cm}}
        \toprule
        Highlighted output sentence & Decontexutalized Fact \\
        \midrule
        \textbf{The} Los Angeles County Fire Department responded to multiple \underline{911} \textbf{calls around 4:30 p.m.} at Penn Park, where the tree had toppled, trapping up to 20 people beneath its branches. & The \underline{911} calls were made around 4:30. \\
        \midrule
        The confirmation hearings for Supreme Court nominee \underline{Brett Kavanaugh} \ldots{} \textbf{Key issues included his views on presidential power}, abortion rights, and potential conflicts of interest regarding the Russia investigation. & Key issues included \underline{Brett Kavanaugh's} views on presidential power.	\\
        \bottomrule
    \end{tabular}
    }
    \caption{Examples illustrating our decontextualization step, drawn from \citet{gunjal-durrett-2024-molecular}. Initially, \ourmethod{} highlights (\textbf{bold}) are reformulated into decontextualized facts ($\rightarrow$). These facts are subsequently aligned with revised highlights ($\leftarrow$, \underline{underlined}), to allow sentence-level attribution to incorporate additional context when needed. For example, in the second row, the mention of \emph{``Brett Kavanaugh''} originates from a separate sentence, requiring the inclusion of additional source text to ensure accurate attribution.}
    \label{tab:decontextualization_examples}
\end{table*}

\subsection{Generating a Decontextualized Output Statement}
As described in \cref{sec:IOLA_task_definition}, a user's query consists of contextualized spans extracted from the output that depend on the surrounding text for full comprehension (e.g., the word \emph{``consumers''} in \cref{fig:iola_example}). 
Step (A) of our method reformulates the selected spans into a self-contained \textit{decontextualized} sentence, for which source attribution can be more easily sought in an unambiguous manner. We use the approach from \citet{gunjal-durrett-2024-molecular}, as exemplified in \cref{tab:decontextualization_examples}.

This process may incorporate in the decontextualized statement additional phrases from the generated output text, beyond the user’s initial highlights. For example, replacing the ambiguous ``they'' pronoun with the explicit ``consumers'' mention in \cref{fig:contributions}, highlighted orange. 
Consequently, the obtained decontextualized statement includes all the information for which attribution should be identified within the source texts.
If the query remains contextualized, this key information may be omitted, resulting in inaccurate attribution. 
By including the additional output span, the attribution used for the first sentence would be included, ensuring comprehensive coverage of the relevant content. For more details, see \cref{sec:alignment_factscore_to_spans}.

\subsection{Query-focused Attribution}\label{sec:laq_methods}
Step (B) of our \ourmethod{} method attributes the decontextualized sentence to the source texts, ensuring factual consistency while minimizing the retrieval of irrelevant spans.
The effectiveness of this step depends on the generation method, particularly whether attribution metadata is available. 
Sentence-level attribution approaches, which provide fixed links between source and output spans, significantly reduce the search space, facilitating the localization of supporting evidence.
In contrast, for non-attributed generation, the system must search the entire source document, increasing computational complexity.

For this step, we explore two approaches: one uses an LLM prompt while the other leverages the model’s internal representations to identify alignments based on hidden state similarities \citep{phukan-etal-2024-peering}.

\paragraph{LLM-based Prompt Alignments.}
Leveraging the strong few-shot learning and reasoning capabilities of LLMs, we prompt an LLM to output the aligned spans. The prompt is listed in \cref{fig:llm_based_alignment_prompt}. Attributed source spans are separated by a semicolon (;). If a span does not match the source text, we apply a fuzzy search.\footnote{\url{https://github.com/google/diff-match-patch}} If the fuzzy search fails, we retry the prompt up to five times. If that also fails, we fall back to the original attribution provided by the attribution metadata, if available. Otherwise, we use all the source documents for attribution.

\paragraph{LLM-based Internals Alignments.}
Another strategy for achieving granular attribution is to compute the cosine similarity between the contextual hidden state representations of the source tokens and the output tokens \citep{dou2021word, phukan-etal-2024-peering}. \citet{phukan-etal-2024-peering} has been shown to surpass GPT-4-based prompting methods in terms of accuracy, but was only evaluated in paragraph-level citations. In this work, we investigate its usefulness in \ourmethod{} settings.\footnote{We re-implemented \citet{phukan-etal-2024-peering}, as no source code was available.} Compared to the previous LLM prompt-based approach, this method requires direct access to the model's weights, necessitating the use of open models.\footnote{For more details on both approaches, see \cref{sec:laq_methods_details}.}

\begin{table*}[t]
    \small
    \centering
    \setlength{\tabcolsep}{0.5em}
    \renewcommand{\arraystretch}{1.2}
    \adjustbox{max width=\textwidth}{
    \begin{tabular}{p{10cm}p{5cm}}
        \toprule
        Output sentence & Example decomposed fact \\
        \midrule
        \textbf{Exposing students to texts from different religions} can be beneficial for their learning, as it helps them understand the development and advancement of societies, \textbf{promoting understanding}, respect, and fellowship. & Exposing students to texts from different religions promotes understanding.\\
        \midrule
        \textbf{Guns} are rarely used in self-defense, are frequently stolen and used by criminals, and their \textbf{presence makes conflicts more likely to become violent}; armed civilians are unlikely to stop crimes and may make situations more deadly. & The presence of gun make conflict more likely to become violent.\\

        \bottomrule
    \end{tabular}
    }
    \caption{Example synthesized \ourmethod{} inputs, simulating a user highlighting the output. First, output sentences are decomposed into atomic facts ($\rightarrow$). Then, these facts are aligned back to highlights ($\leftarrow$), denoted in \textbf{bold}.}
    \label{tab:alignment_of_decomposed_facts}
\end{table*}

\section{Experimental Setup} \label{sec:benchmark}
We evaluate the efficacy of our proposed framework in \cref{sec:frameworks} by benchmarking multiple baseline methods for each stage in the process. We design an experimental setup that assesses both the quality of generated outputs and the accuracy of their attributions. Our evaluation consists of automatic assessments on two key content-grounded generation tasks: Multi-Document Summarization (MDS) and Long-Form Question Answering (LFQA).

This section provides the foundation for benchmarking \ourmethod{} and examining its effectiveness in reducing cognitive load while preserving factual consistency. We first introduce the datasets used in our experiments and describe the methodology for synthesizing attribution queries to simulate user fact-checking behavior (\cref{sec:datasets}). Then, we describe our evaluation framework (\cref{sec:evaluation}), which measures the quality of the attribution under contextualized and decontextualized conditions.

\begin{figure}[t]
    \vspace{-1em}
    \centering
    \includegraphics[width=0.9\linewidth]{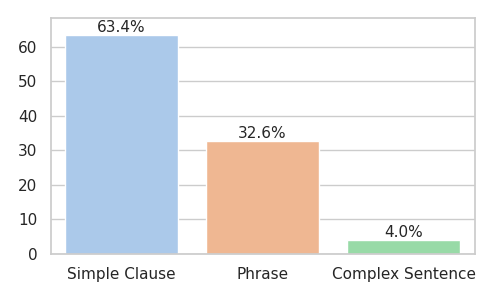}
    \caption{Distribution of span types based on syntactic complexity.}
    \label{fig:spans}
\end{figure}

\subsection{Datasets}\label{sec:datasets}
Our benchmark includes both a multi-document summarization setting (MDS) and a long-form QA setting setting (LFQA). Both are content-grounded settings such that source documents are used to generate an output. Specifically, we use SPARK \citep{ernst-etal-2024-power} for MDS,\footnote{SPARK is a subset of Multi-News \citep{fabbri-etal-2019-multi}} and the RAG-based dataset curated by \citet{liu-etal-2023-evaluating} for LFQA.\footnote{Statistics and more details are provided in \cref{sec:experimental_setup_details}.}

\paragraph{Synthesizing \ourmethod{} Highlights for a Given Output.}

The source documents are used to generate outputs with attribution metadata, as described in \cref{sec:generation_methods}. Given the outputs generated, we synthesize \ourmethod{} inputs by simulating the user's process of highlighting relevant spans.

Our approach for generating highlights involves first decomposing each output sentence into atomic facts and then aligning these facts with the output sentence, exemplified in \cref{tab:alignment_of_decomposed_facts}.
To ensure our decomposition method closely mimics how users select contextualized spans, we adopt the contextualized decomposition approach from FActScore, which was specifically designed to break down long-form generations into atomic facts \citep{min-etal-2023-factscore}. We use GPT-4o \citep{openai2024gpt4ocard} for the decomposition. In order to align the generated output facts with the output, we use a naive lexical-based algorithm, described in \cref{sec:alignment_factscore_to_spans}.

Our process for synthesizing facts results in an average of 2.6 facts per sentence. For each instance, we sample ten facts extracted from the entire output. We report the distribution of facts according to their syntactic complexity as a measure of how diverse the generated facts are. Specifically, we use the following categories:
\begin{enumerate}
    \item \textbf{Phrase}: A span consisting of syntactic constituents without a complete clause structure (i.e., no finite verb or predicate). Example spans include: \textit{``Kavanaugh past writings''}, \textit{``A technical glitch''}.
    
    \item \textbf{Simple Clause}: Contains at least one finite verb. Example spans: \textit{``Judge Brett Kavanaugh faced intense scrutiny''}, \textit{``His previous escape occurred in 2005''}.
    
    \item \textbf{Complex Sentence}: Contains at least one embedded or subordinate clause and explicit discourse connectives. Example: \textit{``Peach trees should be planted while they are dormant''}.
\end{enumerate}

As illustrated in \cref{fig:spans}, the majority of extracted facts are simple clauses (roughly two-thirds), followed by phrases (about one-third), with complex sentences making up only a small proportion.\footnote{Categorization was performed using the SpaCy NLP toolkit: \url{https://spacy.io/}}

\subsection{Generation Baselines}\label{sec:generation_methods}
Following our suggested framework in \cref{sec:frameworks}, we benchmark three baseline methods for the first generation stage, from methods that provide no attribution to those that provide fine-grained attribution. Full details for the following methods are provided in \cref{sec:generation_methods_details}.

\paragraph{Vanilla.} We include a naive baseline that generates text without attribution, as this represents a common approach in many real-world applications where attribution is not explicitly modeled. Evaluating this baseline allows us to measure the extent to which \ourmethod{} methods can provide correct attribution on the entire source documents.

\paragraph{ALCE.} \citet{gao-etal-2023-enabling} is a prominent attribution method that prompts the LLM to add citations at the end of each output sentence, in the form of square bracket, such as ``\ldots{}[1].'' This method provides a fairly coarse-grained attribution, as citations point to an entire source document.

\paragraph{Attr. First.} \citet{slobodkin-etal-2024-attribute} divide the generation process into multiple explicit steps, allowing the attribution to be traced back to source spans. 
The first step, content selection, involves highlighting relevant source spans. 
The generation is then constrained to these selected spans, allowing the output to be tied back to the source. 
Unlike ALCE, which attributes at the document level, this approach attributes source spans, significantly reducing the costs associated with \ourmethod{} while increasing the number of tokens required for generating the initial output. 
We analyze this trade-off in \cref{sec:cost_analysis}.

\subsection{Evaluation}\label{sec:evaluation}
Our evaluation is comprised of different metrics for the quality of the output, following standard practices of each task, as well as the quality of the citations, adapted to the \ourmethod{} setting. The purpose of measuring output quality is to show that overall methods that support localized attribution do not hurt output quality with respect to relevance. We incorporate both automated and human evaluations into our methodology.

\paragraph{Automatic evaluation}

To evaluate the quality of the output, we follow \citet{slobodkin-etal-2024-attribute} and calculate Rouge-L \citep{lin-2004-rouge} and BertScore \citep{bert-score}, which were also used in their study. Additionally, we calculate METEOR \citep{banarjee2005} and BLEURT20 \citep{sellam2020bleurt}. Rouge-L and METEOR utilize n-gram comparisons, while BertScore and BLEURT20 are based on language models. All these metrics compare the generated output to a reference output. Lastly, we include a fluency metric based on MAUVE \citep{pillutla-etal:mauve:neurips2021}, which compares the distribution of the output to that of the reference texts.

To evaluate \ourmethod{} citations, we sample ten facts from the facts extracted from the output, as described in \cref{sec:datasets}. We then calculate AutoAIS \citep{gao-etal-2023-rarr}, which is an entailment metric commonly used for evaluating attribution. The metric outputs binary classification of whether an attributed source text supports an output fact, which is then averaged across all output facts to calculate the final score.
Following \citet{gao-etal-2023-rarr}, we make the distinction between evaluating entailment with contextualized facts and decontextualized facts. %
We source contextualized facts from the process described in \cref{sec:datasets}, and decontextualized facts from the process described in \cref{sec:laq_methods}.

Additionally, we measure the attributed text length in content words\footnote{Excluding stop-words \url{https://nltk.org}} to confirm that our method significantly reduces unnecessary reading.
Lastly, we report the percent of non-attributed facts.\footnote{More details are provided in \cref{sec:evaluation_details}.}

\begin{table*}[h!]
    \centering
    \small
    \adjustbox{max width=\linewidth}{
        \begin{tabular}{llccccc}
            \toprule
             & Method & \textsc{R-L $\uparrow$} & \textsc{METEOR $\uparrow$} & \textsc{BertScore $\uparrow$} & \textsc{BLEURT-20 $\uparrow$} & MAUVE $\uparrow$\\
             \midrule
             \parbox[t]{2mm}{\multirow{3}{*}{\rotatebox[origin=c]{90}{MDS}}} & \textsc{Vanilla} & 19.2 \tiny{$\pm{0.6}$} & 28.3 & 86.4 \tiny{$\pm{0.2}$} & \textbf{43.0} \tiny{$\pm{0.7}$} & 59.8 \\
             & \textsc{ALCE} & 19.4 \tiny{$\pm{0.6}$} & 27.3 & 86.1 \tiny{$\pm{0.2}$} & 38.2 \tiny{$\pm{0.8}$} & 63.7 \\
             & \textsc{Attr. First} & \textbf{21.1} \tiny{$\pm{0.7}$} & \textbf{29.7} & \textbf{86.6} \tiny{$\pm{0.2}$} & 41.1 \tiny{$\pm{0.9}$} & \textbf{84.9}  \\
             \toprule
             \parbox[t]{2mm}{\multirow{3}{*}{\rotatebox[origin=c]{90}{LFQA}}} & \textsc{Vanilla} & 37.2 \tiny{$\pm{3.2}$} & 45.6 & \textbf{90.7} \tiny{$\pm{0.6}$} & \textbf{60.5} \tiny{$\pm{1.7}$} & 81.5 \\
             & \textsc{ALCE} & 34.4 \tiny{$\pm{2.7}$} & 44.3 & 90.1 \tiny{$\pm{0.5}$} & 56.8 \tiny{$\pm{1.7}$} & 90.6 \\
             & \textsc{Attr. First} & \textbf{38.2} \tiny{$\pm{2.7}$} & \textbf{46.1} & 90.6 \tiny{$\pm{0.6}$} & 58.5 \tiny{$\pm{1.8}$} & \textbf{96.7} \\
             \bottomrule
        \end{tabular}
    }
    \caption{Generated text quality results, averages include standard error of the mean.}
    \label{tab:output_quality_results}
\end{table*}

\section{Results and Analyses}\label{sec:Results_and_Analyses}

\subsection{Main Results}
\eh{Thoughts on diff: the generating algorithm can completely change the upper bound on attribution quality, makes more sense to report diffs}
The output quality metrics are reported in \cref{tab:output_quality_results}. \textsc{Attr. First} outperforms other methods in terms of \textsc{Rouge-L}, \textsc{METEOR} and \textsc{MAUVE}, while the Vanilla generation outperforms in terms of \textsc{BLEURT-20}. This suggests that \textsc{Attr. First} has more lexical overlap with the reference outputs, while the Vanilla generation is more semantically similar. In general, both methods achieve similar output quality results, with ALCE lagging behind.

\begin{table*}[t]
    \centering
    \small
    \adjustbox{max width=\textwidth}{
        \begin{tabular}{llcccc}
            \toprule
            \parbox[t]{3mm}{\multirow{9}{*}{\rotatebox[origin=c]{90}{MDS \qquad{}}}} &  Method & \textsc{AutoAIS Con. $\uparrow$} & \textsc{AutoAIS Decon. $\uparrow$} & \textsc{Length $\downarrow$} & \textsc{Non Att. (\%)  $\downarrow$} \\
             \midrule
             & \textsc{Vanilla} & \textbf{82.2} \tiny{$\pm{1.6}$} & \textbf{84.5} \tiny{$\pm{2.0}$} & 1681.6 \tiny{$\pm{205.5}$} & 0.0 \\
             & \multirow{2}{*}{\yellowmarker}\quad{} LLM Prompt & 62.5 \tiny{$\pm{2.0}$} & \myyellow{49.7} \tiny{$\pm{2.5}$} & 32.0 \tiny{$\pm{1.8}$} & 0.0 \\
             & \qquad LLM Internals & 18.0 \tiny{$\pm{1.7}$} & 13.1 \tiny{$\pm{1.5}$} & 28.1 \tiny{$\pm{0.9}$} & 0.0 \\
             \cmidrule{2-6}
             & \textsc{ALCE} & 67.4 \tiny{$\pm{2.3}$} & 74.8 \tiny{$\pm{2.3}$} & 979.1 \tiny{$\pm{117.8}$} & 5.2 \tiny{$\pm{0.8}$} \\
             & \multirow{2}{*}{\yellowmarker}\quad{} LLM Prompt & 55.8 \tiny{$\pm{2.2}$} & 44.3 \tiny{$\pm{2.4}$} & 41.6 \tiny{$\pm{3.4}$} & 5.2 \tiny{$\pm{0.8}$} \\
             & \qquad LLM Internals & 15.5 \tiny{$\pm{1.6}$} & 10.2 \tiny{$\pm{1.5}$} & 29.9 \tiny{$\pm{8.2}$} & 8.2 \tiny{$\pm{1.2}$} \\
             \cmidrule{2-6}
             & \textsc{Attr. First} & 80.3 \tiny{$\pm{2.2}$} & 58.0 \tiny{$\pm{2.8}$} & 33.0 \tiny{$\pm{2.4}$} & 0.4 \tiny{$\pm{0.2}$} \\
             &  \multirow{2}{*}{\yellowmarker}\quad{} LLM Prompt & \myyellow{71.5} \tiny{$\pm{2.3}$} & 42.4 \tiny{$\pm{2.4}$} & 14.6 \tiny{$\pm{0.5}$} & 0.4 \tiny{$\pm{0.2}$} \\
             & \qquad LLM Internals & 28.6 \tiny{$\pm{2.4}$} & 13.2 \tiny{$\pm{1.7}$} & \textbf{12.2} \tiny{$\pm{0.4}$} & 21.4 \tiny{$\pm{0.9}$} \\
             \toprule
             \parbox[t]{3mm}{\multirow{9}{*}{\rotatebox[origin=c]{90}{LFQA}}} & \textsc{Vanilla} & 69.5 \tiny{$\pm{4.6}$} & 71.0 \tiny{$\pm{4.5}$} & 4636.8 \tiny{$\pm{488.3}$} & 0.0 \\
             &  \multirow{2}{*}{\yellowmarker}\quad{} LLM Prompt & 65.1 \tiny{$\pm{3.8}$} & 65.4 \tiny{$\pm{4.3}$} & 38.1 \tiny{$\pm{2.6}$} & 0.0 \\
             & \qquad LLM Internals & 19.0 \tiny{$\pm{2.8}$} & 18.0 \tiny{$\pm{2.4}$} & 24.9 \tiny{$\pm{1.5}$} & 0.0 \\
             \cmidrule{2-6}
             & \textsc{ALCE} & 50.8 \tiny{$\pm{4.8}$} & 55.6 \tiny{$\pm{5.1}$} & 2346.0 \tiny{$\pm{300.2}$} & 13.8 \tiny{$\pm{4.2}$} \\
             &  \multirow{2}{*}{\yellowmarker}\quad{} LLM Prompt & 56.8 \tiny{$\pm{4.0}$} & 52.8 \tiny{$\pm{3.9}$} & 42.0 \tiny{$\pm{10.9}$} & 13.8 \tiny{$\pm{4.2}$} \\
             & \qquad LLM Internals & 13.0 \tiny{$\pm{2.4}$} & 12.8 \tiny{$\pm{2.4}$} & 26.6 \tiny{$\pm{1.5}$} & 17.1 \tiny{$\pm{3.0}$} \\
             \cmidrule{2-6}
             & \textsc{Attr. First} & \textbf{88.0} \tiny{$\pm{3.4}$} & \textbf{83.9} \tiny{$\pm{3.3}$} & 43.3 \tiny{$\pm{2.4}$} & 0.0 \\
             &  \multirow{2}{*}{\yellowmarker}\quad{} LLM Prompt & \myyellow{83.0} \tiny{$\pm{3.1}$} & \myyellow{69.6} \tiny{$\pm{4.3}$} & 17.3 \tiny{$\pm{0.8}$} & 0.0 \\
             & \qquad LLM Internals & 46.6 \tiny{$\pm{4.0}$} & 37.8 \tiny{$\pm{4.4}$} & \textbf{14.3} \tiny{$\pm{0.7}$} & 7.0 \tiny{$\pm{1.7}$} \\
             \bottomrule
        \end{tabular}
    }
    \caption{\ourmethod{} citation results, averages include standard error of the mean. We separately calculate AutoAIS for contextualizd (Con.) and decontextualized (Decon.) output facts. \yellowmarker{} indicates \ourmethod{} methods and \myyellow{yellow} indicates the best \ourmethod{} method. Non Attributed measures the percentage of facts without attribution. }
    \label{tab:citation_quality_results}
\end{table*}

\begin{table}[t]
    \centering
    \small
    \adjustbox{max width=\textwidth}{
        \begin{tabular}{llc}
            \toprule
            \parbox[t]{3mm}{\multirow{4}{*}{\rotatebox[origin=c]{90}{MDS \qquad{}}}} &  Method & \textsc{AIS Decon. $\uparrow$} \\
             \midrule
             & \textsc{Vanilla} & \textbf{91.5} \tiny{$\pm{2.3}$} \\
             & \multirow{1}{*}{\yellowmarker}\quad{} LLM Prompt & \myyellow{39.0} \tiny{$\pm{4.2}$} \\
             \cmidrule{2-3}
             & \textsc{Attr. First} & 54.3 \tiny{$\pm{4.4}$}  \\
             &  \multirow{1}{*}{\yellowmarker}\quad{} LLM Prompt & 31.6 \tiny{$\pm{3.8}$} \\
             \toprule
             \parbox[t]{3mm}{\multirow{4}{*}{\rotatebox[origin=c]{90}{LFQA}}} & \textsc{Vanilla} & \textbf{90.9} \tiny{$\pm{5.1}$} \\
             &  \multirow{1}{*}{\yellowmarker}\quad{} LLM Prompt & \myyellow{59.5} \tiny{$\pm{7.7}$} \\
             \cmidrule{2-3}
             & \textsc{Attr. First} & 53.6 \tiny{$\pm{8.6}$} \\
             &  \multirow{1}{*}{\yellowmarker}\quad{} LLM Prompt & 50.2 \tiny{$\pm{9.2}$} \\
             \bottomrule
        \end{tabular}
    }
    
    \caption{\ourmethod{} human evaluation of citation results, averages include standard error of the mean. \yellowmarker{} indicates \ourmethod{} methods.}
    \label{tab:human_eval_results}
\end{table}

The citations quality metrics are reported in \cref{tab:citation_quality_results}. We make the following observations.

\paragraph{\ourmethod{} methods significantly and attractively reduce the length of the attributed text.}
Across all methods, \ourmethod{} reduces attribution length by two orders of magnitude for Vanilla and ALCE, and by an average of 59\% for \textsc{Attr. First}.
For example, in the Vanilla setting, which does not rely on a particular generation method but does not provide any attribution, \ourmethod{} attribution can direct the user to correct highly localized supporting spans in nearly two thirds of the cases.

\paragraph{The LLM prompt is the best-performing \ourmethod{} method.} In all generation methods, we find that the LLM prompt performs the best in terms of AutoAIS, significantly surpassing the LLM internals method. This is true for both MDS and LFQA settings.
The LLM internals method has low performance across all generation methods. The best results for the LLM internals are achieved when the source is localized with \textsc{Attr. First}, suggesting that it struggles with localization of document-level texts. In addition, when the LLM internals method is applied on top of source-localized attribution methods, ALCE and \textsc{Attr. First}, we observe an increase in non-attributed output words.

\paragraph{\ourmethod{} methods can leverage localized attributions provided by \textsc{Attr. First}.}
Even without applying \ourmethod{}, \textsc{Attr. First} provides very concise sentence-level attribution, averaging only 36 characters. This means that the localized support for the \ourmethod{} fact needs to be identified only within a quite short span. Consequently, the strong performance of \textsc{Attr. First} carries over to \ourmethod{}. In the contextualized setting, \textsc{Attr. First} is the top-performing \ourmethod{} method, indicating that \ourmethod{} methods can leverage initially localized attributions provided by the generation method itself.
However, in the decontextualized setting, \textsc{Attr. First} yields notably low AutoAIS scores, as low as 58 for MDS, and 53.6 for LFQA in our manual evaluation, described in \cref{sec:human_analysis}. These low scores limit the effectiveness of \ourmethod{} methods, since the necessary evidence for the decontextualized facts is absent from the original \textsc{Attr. First} provided spans.
We hypothesize that this degredation stems from \textsc{Attr. First} failure to decontextualize its attributions. This suggests that when generating attributions for localized output segments, it is crucial to first decontextualize these output spans, and accordingly to make sure to support also the decontextualizing information within the source attributions. %

\subsection{Human Analysis}\label{sec:human_analysis}

To further assess our findings, we report a small-scale human annotation conducted by the authors using our most promising methods. We annotated 20 examples per task, each for the Vanilla and \textsc{Attr. First} methods, both with and without LAQuer, resulting in 80 examples per task (160 in total). For each example, we calculate AIS \citep{rashkin-etal-2023-measuring} at the decontextualized fact-level. For AIS, similar to the AutoAIS metric, the annotator is asked to make a binary classification of whether an output fact is supported by the attributed source texts; we then average classifications across all output facts to calculate the final score.

Our results are reported in \cref{tab:human_eval_results}. In accordance with our main results in \cref{tab:citation_quality_results}, we find that \ourmethod{} methods struggle with decontextualized facts.
From this analysis, we observe that the model often omits the document's broader theme. For example, in \cref{tab:results_localization_examples_e2e}, the LLM prompt method correctly attributes multiple ``issues'', yet it fails to attribute ``Supreme Court''.

\eh{consider calculating also correlation with AutoAIS}

\subsection{Cost Analysis}\label{sec:cost_analysis}
We provide the average size of prompts in \cref{tab:size_of_prompts}. On one hand, we find that \ourmethod{} prompts in \textsc{Attr. First} are an order of magnitude smaller than in Vanilla generation. On the other hand, \textsc{Attr. First} generation is costly, inducing an increase of 90\% in prompt length compared to Vanilla generation, as reported by \citet{slobodkin-etal-2024-attribute}. These results suggest that increased computational cost during generation can lead to more efficient \ourmethod{} methods.

\subsection{Estimate for \ourmethod{} Localization}
To better understand the potential benefits of \ourmethod{}, we estimate the average amount of text required to support an output fact or sentence. We compare this across different levels of source granularity, including source spans, source sentences, and entire source documents. For this analysis, we utilize the SPARK dataset \citep{ernst-etal-2024-power}, which is used in our study and contains fine-grained, human-annotated attribution.
 
\begin{table}[h!]
    \small
    \centering
    \setlength{\tabcolsep}{0.5em}
    \adjustbox{max width=\textwidth}{
    \begin{tabular}{l|cc}
    \toprule
    Source granularity & Output facts & Output sentences\\
    \midrule
    Spans & 128.0 & 231.4\\
    Sentence & 278.5 & 485.1\\
    Document & 4679.5  & 7226.6 \\
    \bottomrule
    \end{tabular}
    }
    \caption{Analysis of attribution lengths (measured in characters) with varying granularities, based on the SPARK dataset \citep{ernst-etal-2024-power}.}
    \label{tab:spark_analysis}
\end{table}

Our analysis, summarized in \cref{tab:spark_analysis}, presents the average number of characters to read under different attribution granularities. \ourmethod{} operates at both the source and output fact levels, requiring an average of 128 characters to read. In contrast, \textsc{Attr. First} attributes at the output sentence level with source spans, resulting in an average of 278.5 characters. This finding highlights the benefits of localizing attribution per output fact, reducing the text users need to read by 54\%.

\section{Conclusion}\label{sec:Discussion}
\eh{Consider showing performance segmented by span type}

In this work, we introduce a novel motivation for post-hoc attributed text generation, enabling users to create localized attribution queries, \ourmethod{}. We introduce a challenging benchmark, which subsumes existing attribution methods by considering both the generation and post-hoc steps.
Our results show that \ourmethod{} methods significantly reduce attribution length, but \ourmethod{} attribution remains a challenging task for decontextualized facts. In addition, our methods are associated with a high cost of LLM calls, suggesting future research should focus on creating more efficient frameworks. Lastly, there is a performance gap between different generation methods.

\section*{Limitations}

Addressing attribution queries increases computational cost on top of fixed sentence-level or token-level attribution. In \cref{sec:cost_analysis}, we discuss the trade-off between computational cost during generation and that during attribution.

While our work is focused on content-grounded generation, \ourmethod{} could be applied to outputs generated by the model's parametric knowledge, by retrieving the documents after the generation rather than before. We leave such exploration for future work.

AutoAIS is used as a key metric for evaluating attribution quality, which is an  LLM-based automated metric. We conducted a small-scale human analysis to support these results in \cref{sec:human_analysis}, finding similar trends.%

\section*{Ethical Considerations}
The ability to attribute outputs of LLMs to specific sources is crucial for transparency, accountability, and trust in AI-generated content. Our work contributes to this goal by simplifying the attribution process for users and making it more localized. However, errors in attribution can mislead users into assuming a stronger or weaker connection between the generated content and its source than what actually exists.

We utilized AI-assisted writing tools during the preparation of this paper to improve clarity and coherence. However, all content was carefully reviewed and edited by the authors to ensure accuracy.

\section*{Acknowledgements}
We would like to thank our reviewers for their constructive suggestions and comments. This work was supported by the Israel Science Foundation (grant no. 2827/21), NSF-CAREER Award 1846185, and NSF-AI Engage Institute DRL-2112635.

\bibliography{custom}

\begin{thebibliography}{30}
\providecommand{\natexlab}[1]{#1}

\bibitem[{Banerjee and Lavie(2005)}]{banarjee2005}
Satanjeev Banerjee and Alon Lavie. 2005.
\newblock {METEOR}: An automatic metric for {MT} evaluation with improved correlation with human judgments.
\newblock In \emph{Proceedings of the {ACL} Workshop on Intrinsic and Extrinsic Evaluation Measures for Machine Translation and/or Summarization}, pages 65--72, Ann Arbor, Michigan. Association for Computational Linguistics.

\bibitem[{Bohnet et~al.(2023)Bohnet, Tran, Verga, Aharoni, Andor, Soares, Ciaramita, Eisenstein, Ganchev, Herzig, Hui, Kwiatkowski, Ma, Ni, Saralegui, Schuster, Cohen, Collins, Das, Metzler, Petrov, and Webster}]{bohnet2023attributed}
Bernd Bohnet, Vinh~Q. Tran, Pat Verga, Roee Aharoni, Daniel Andor, Livio~Baldini Soares, Massimiliano Ciaramita, Jacob Eisenstein, Kuzman Ganchev, Jonathan Herzig, Kai Hui, Tom Kwiatkowski, Ji~Ma, Jianmo Ni, Lierni~Sestorain Saralegui, Tal Schuster, William~W. Cohen, Michael Collins, Dipanjan Das, Donald Metzler, Slav Petrov, and Kellie Webster. 2023.
\newblock \href {https://arxiv.org/abs/2212.08037} {Attributed question answering: Evaluation and modeling for attributed large language models}.
\newblock \emph{Preprint}, arXiv:2212.08037.

\bibitem[{Buchmann et~al.(2024)Buchmann, Liu, and Gurevych}]{buchmann-etal-2024-attribute}
Jan Buchmann, Xiao Liu, and Iryna Gurevych. 2024.
\newblock \href {https://doi.org/10.18653/v1/2024.emnlp-main.463} {Attribute or abstain: Large language models as long document assistants}.
\newblock In \emph{Proceedings of the 2024 Conference on Empirical Methods in Natural Language Processing}, pages 8113--8140, Miami, Florida, USA. Association for Computational Linguistics.

\bibitem[{Choi et~al.(2021)Choi, Palomaki, Lamm, Kwiatkowski, Das, and Collins}]{choi-etal-2021-decontextualization}
Eunsol Choi, Jennimaria Palomaki, Matthew Lamm, Tom Kwiatkowski, Dipanjan Das, and Michael Collins. 2021.
\newblock \href {https://doi.org/10.1162/tacl_a_00377} {Decontextualization: Making sentences stand-alone}.
\newblock \emph{Transactions of the Association for Computational Linguistics}, 9:447--461.

\bibitem[{Cohen-Wang et~al.(2024)Cohen-Wang, Shah, Georgiev, and Madry}]{cohen-wang2024contextcite}
Benjamin Cohen-Wang, Harshay Shah, Kristian Georgiev, and Aleksander Madry. 2024.
\newblock \href {https://openreview.net/forum?id=7CMNSqsZJt} {Contextcite: Attributing model generation to context}.
\newblock In \emph{The Thirty-eighth Annual Conference on Neural Information Processing Systems}.

\bibitem[{Ding et~al.(2024)Ding, Luo, Cao, and Luo}]{ding2024attentiondependencyparsingaugmentation}
Qiang Ding, Lvzhou Luo, Yixuan Cao, and Ping Luo. 2024.
\newblock \href {https://arxiv.org/abs/2412.11404} {Attention with dependency parsing augmentation for fine-grained attribution}.
\newblock \emph{Preprint}, arXiv:2412.11404.

\bibitem[{Dou and Neubig(2021)}]{dou2021word}
Zi-Yi Dou and Graham Neubig. 2021.
\newblock Word alignment by fine-tuning embeddings on parallel corpora.
\newblock In \emph{Conference of the European Chapter of the Association for Computational Linguistics (EACL)}.

\bibitem[{Ernst et~al.(2024)Ernst, Shapira, Slobodkin, Adar, Bansal, Goldberger, Levy, and Dagan}]{ernst-etal-2024-power}
Ori Ernst, Ori Shapira, Aviv Slobodkin, Sharon Adar, Mohit Bansal, Jacob Goldberger, Ran Levy, and Ido Dagan. 2024.
\newblock \href {https://doi.org/10.18653/v1/2024.findings-acl.389} {The power of summary-source alignments}.
\newblock In \emph{Findings of the Association for Computational Linguistics: ACL 2024}, pages 6527--6548, Bangkok, Thailand. Association for Computational Linguistics.

\bibitem[{Fabbri et~al.(2019)Fabbri, Li, She, Li, and Radev}]{fabbri-etal-2019-multi}
Alexander Fabbri, Irene Li, Tianwei She, Suyi Li, and Dragomir Radev. 2019.
\newblock \href {https://doi.org/10.18653/v1/P19-1102} {Multi-news: A large-scale multi-document summarization dataset and abstractive hierarchical model}.
\newblock In \emph{Proceedings of the 57th Annual Meeting of the Association for Computational Linguistics}, pages 1074--1084, Florence, Italy. Association for Computational Linguistics.

\bibitem[{Gao et~al.(2023{\natexlab{a}})Gao, Dai, Pasupat, Chen, Chaganty, Fan, Zhao, Lao, Lee, Juan, and Guu}]{gao-etal-2023-rarr}
Luyu Gao, Zhuyun Dai, Panupong Pasupat, Anthony Chen, Arun~Tejasvi Chaganty, Yicheng Fan, Vincent Zhao, Ni~Lao, Hongrae Lee, Da-Cheng Juan, and Kelvin Guu. 2023{\natexlab{a}}.
\newblock \href {https://doi.org/10.18653/v1/2023.acl-long.910} {{RARR}: Researching and revising what language models say, using language models}.
\newblock In \emph{Proceedings of the 61st Annual Meeting of the Association for Computational Linguistics (Volume 1: Long Papers)}, pages 16477--16508, Toronto, Canada. Association for Computational Linguistics.

\bibitem[{Gao et~al.(2023{\natexlab{b}})Gao, Yen, Yu, and Chen}]{gao-etal-2023-enabling}
Tianyu Gao, Howard Yen, Jiatong Yu, and Danqi Chen. 2023{\natexlab{b}}.
\newblock \href {https://doi.org/10.18653/v1/2023.emnlp-main.398} {Enabling large language models to generate text with citations}.
\newblock In \emph{Proceedings of the 2023 Conference on Empirical Methods in Natural Language Processing}, pages 6465--6488, Singapore. Association for Computational Linguistics.

\bibitem[{Gunjal and Durrett(2024)}]{gunjal-durrett-2024-molecular}
Anisha Gunjal and Greg Durrett. 2024.
\newblock \href {https://doi.org/10.18653/v1/2024.findings-emnlp.215} {Molecular facts: Desiderata for decontextualization in {LLM} fact verification}.
\newblock In \emph{Findings of the Association for Computational Linguistics: EMNLP 2024}, pages 3751--3768, Miami, Florida, USA. Association for Computational Linguistics.

\bibitem[{Honovich et~al.(2022)Honovich, Aharoni, Herzig, Taitelbaum, Kukliansy, Cohen, Scialom, Szpektor, Hassidim, and Matias}]{honovich-etal-2022-true-evaluating}
Or~Honovich, Roee Aharoni, Jonathan Herzig, Hagai Taitelbaum, Doron Kukliansy, Vered Cohen, Thomas Scialom, Idan Szpektor, Avinatan Hassidim, and Yossi Matias. 2022.
\newblock \href {https://doi.org/10.18653/v1/2022.naacl-main.287} {{TRUE}: Re-evaluating factual consistency evaluation}.
\newblock In \emph{Proceedings of the 2022 Conference of the North American Chapter of the Association for Computational Linguistics: Human Language Technologies}, pages 3905--3920, Seattle, United States. Association for Computational Linguistics.

\bibitem[{Lewis et~al.(2020)Lewis, Perez, Piktus, Petroni, Karpukhin, Goyal, K\"{u}ttler, Lewis, Yih, Rockt\"{a}schel, Riedel, and Kiela}]{RAG}
Patrick Lewis, Ethan Perez, Aleksandra Piktus, Fabio Petroni, Vladimir Karpukhin, Naman Goyal, Heinrich K\"{u}ttler, Mike Lewis, Wen-tau Yih, Tim Rockt\"{a}schel, Sebastian Riedel, and Douwe Kiela. 2020.
\newblock \href {https://proceedings.neurips.cc/paper_files/paper/2020/file/6b493230205f780e1bc26945df7481e5-Paper.pdf} {Retrieval-augmented generation for knowledge-intensive nlp tasks}.
\newblock In \emph{Advances in Neural Information Processing Systems}, volume~33, pages 9459--9474. Curran Associates, Inc.

\bibitem[{Lin(2004)}]{lin-2004-rouge}
Chin-Yew Lin. 2004.
\newblock \href {https://aclanthology.org/W04-1013/} {{ROUGE}: A package for automatic evaluation of summaries}.
\newblock In \emph{Text Summarization Branches Out}, pages 74--81, Barcelona, Spain. Association for Computational Linguistics.

\bibitem[{Liu et~al.(2023)Liu, Zhang, and Liang}]{liu-etal-2023-evaluating}
Nelson Liu, Tianyi Zhang, and Percy Liang. 2023.
\newblock \href {https://doi.org/10.18653/v1/2023.findings-emnlp.467} {Evaluating verifiability in generative search engines}.
\newblock In \emph{Findings of the Association for Computational Linguistics: EMNLP 2023}, pages 7001--7025, Singapore. Association for Computational Linguistics.

\bibitem[{Menick et~al.(2022)Menick, Trebacz, Mikulik, Aslanides, Song, Chadwick, Glaese, Young, Campbell-Gillingham, Irving, and McAleese}]{Menick2022TeachingLM}
Jacob Menick, Maja Trebacz, Vladimir Mikulik, John Aslanides, Francis Song, Martin Chadwick, Mia Glaese, Susannah Young, Lucy Campbell-Gillingham, Geoffrey Irving, and Nat McAleese. 2022.
\newblock \href {https://api.semanticscholar.org/CorpusID:247594830} {Teaching language models to support answers with verified quotes}.
\newblock \emph{ArXiv}, abs/2203.11147.

\bibitem[{Min et~al.(2023)Min, Krishna, Lyu, Lewis, Yih, Koh, Iyyer, Zettlemoyer, and Hajishirzi}]{min-etal-2023-factscore}
Sewon Min, Kalpesh Krishna, Xinxi Lyu, Mike Lewis, Wen-tau Yih, Pang Koh, Mohit Iyyer, Luke Zettlemoyer, and Hannaneh Hajishirzi. 2023.
\newblock \href {https://doi.org/10.18653/v1/2023.emnlp-main.741} {{FA}ct{S}core: Fine-grained atomic evaluation of factual precision in long form text generation}.
\newblock In \emph{Proceedings of the 2023 Conference on Empirical Methods in Natural Language Processing}, pages 12076--12100, Singapore. Association for Computational Linguistics.

\bibitem[{Mishra et~al.(2024)Mishra, Asai, Balachandran, Wang, Neubig, Tsvetkov, and Hajishirzi}]{Mishra2024FinegrainedHD}
Abhika Mishra, Akari Asai, Vidhisha Balachandran, Yizhong Wang, Graham Neubig, Yulia Tsvetkov, and Hannaneh Hajishirzi. 2024.
\newblock \href {https://api.semanticscholar.org/CorpusID:266999558} {Fine-grained hallucination detection and editing for language models}.
\newblock \emph{ArXiv}, abs/2401.06855.

\bibitem[{OpenAI(2024)}]{openai2024gpt4ocard}
OpenAI. 2024.
\newblock \href {https://arxiv.org/abs/2410.21276} {Gpt-4o system card}.
\newblock \emph{Preprint}, arXiv:2410.21276.

\bibitem[{Phukan et~al.(2024)Phukan, Somasundaram, Saxena, Goswami, and Srinivasan}]{phukan-etal-2024-peering}
Anirudh Phukan, Shwetha Somasundaram, Apoorv Saxena, Koustava Goswami, and Balaji~Vasan Srinivasan. 2024.
\newblock \href {https://doi.org/10.18653/v1/2024.findings-acl.682} {Peering into the mind of language models: An approach for attribution in contextual question answering}.
\newblock In \emph{Findings of the Association for Computational Linguistics: ACL 2024}, pages 11481--11495, Bangkok, Thailand. Association for Computational Linguistics.

\bibitem[{Pillutla et~al.(2021)Pillutla, Swayamdipta, Zellers, Thickstun, Welleck, Choi, and Harchaoui}]{pillutla-etal:mauve:neurips2021}
Krishna Pillutla, Swabha Swayamdipta, Rowan Zellers, John Thickstun, Sean Welleck, Yejin Choi, and Zaid Harchaoui. 2021.
\newblock Mauve: Measuring the gap between neural text and human text using divergence frontiers.
\newblock In \emph{NeurIPS}.

\bibitem[{Qi et~al.(2024)Qi, Sarti, Fern{\'a}ndez, and Bisazza}]{qi-etal-2024-model}
Jirui Qi, Gabriele Sarti, Raquel Fern{\'a}ndez, and Arianna Bisazza. 2024.
\newblock \href {https://doi.org/10.18653/v1/2024.emnlp-main.347} {Model internals-based answer attribution for trustworthy retrieval-augmented generation}.
\newblock In \emph{Proceedings of the 2024 Conference on Empirical Methods in Natural Language Processing}, pages 6037--6053, Miami, Florida, USA. Association for Computational Linguistics.

\bibitem[{Ram et~al.(2023)Ram, Levine, Dalmedigos, Muhlgay, Shashua, Leyton-Brown, and Shoham}]{ram-etal-2023-context}
Ori Ram, Yoav Levine, Itay Dalmedigos, Dor Muhlgay, Amnon Shashua, Kevin Leyton-Brown, and Yoav Shoham. 2023.
\newblock \href {https://doi.org/10.1162/tacl_a_00605} {In-context retrieval-augmented language models}.
\newblock \emph{Transactions of the Association for Computational Linguistics}, 11:1316--1331.

\bibitem[{Rashkin et~al.(2023)Rashkin, Nikolaev, Lamm, Aroyo, Collins, Das, Petrov, Tomar, Turc, and Reitter}]{rashkin-etal-2023-measuring}
Hannah Rashkin, Vitaly Nikolaev, Matthew Lamm, Lora Aroyo, Michael Collins, Dipanjan Das, Slav Petrov, Gaurav~Singh Tomar, Iulia Turc, and David Reitter. 2023.
\newblock \href {https://doi.org/10.1162/coli_a_00486} {Measuring attribution in natural language generation models}.
\newblock \emph{Computational Linguistics}, 49(4):777--840.

\bibitem[{Schuster et~al.(2024)Schuster, Lelkes, Sun, Gupta, Berant, Cohen, and Metzler}]{schuster-etal-2024-semqa}
Tal Schuster, Adam Lelkes, Haitian Sun, Jai Gupta, Jonathan Berant, William Cohen, and Donald Metzler. 2024.
\newblock \href {https://doi.org/10.18653/v1/2024.naacl-long.74} {{SEMQA}: Semi-extractive multi-source question answering}.
\newblock In \emph{Proceedings of the 2024 Conference of the North American Chapter of the Association for Computational Linguistics: Human Language Technologies (Volume 1: Long Papers)}, pages 1363--1381, Mexico City, Mexico. Association for Computational Linguistics.

\bibitem[{Sellam et~al.(2020)Sellam, Das, and Parikh}]{sellam2020bleurt}
Thibault Sellam, Dipanjan Das, and Ankur~P Parikh. 2020.
\newblock Bleurt: Learning robust metrics for text generation.
\newblock In \emph{Proceedings of ACL}.

\bibitem[{Slobodkin et~al.(2024)Slobodkin, Hirsch, Cattan, Schuster, and Dagan}]{slobodkin-etal-2024-attribute}
Aviv Slobodkin, Eran Hirsch, Arie Cattan, Tal Schuster, and Ido Dagan. 2024.
\newblock \href {https://doi.org/10.18653/v1/2024.acl-long.182} {Attribute first, then generate: Locally-attributable grounded text generation}.
\newblock In \emph{Proceedings of the 62nd Annual Meeting of the Association for Computational Linguistics (Volume 1: Long Papers)}, pages 3309--3344, Bangkok, Thailand. Association for Computational Linguistics.

\bibitem[{Thoppilan et~al.(2022)Thoppilan, Freitas, Hall, Shazeer, Kulshreshtha, Cheng, Jin, Bos, Baker, Du, Li, Lee, Zheng, Ghafouri, Menegali, Huang, Krikun, Lepikhin, Qin, Chen, Xu, Chen, Roberts, Bosma, Zhou, Chang, Krivokon, Rusch, Pickett, Meier-Hellstern, Morris, Doshi, Santos, Duke, S{\o}raker, Zevenbergen, Prabhakaran, D{\'i}az, Hutchinson, Olson, Molina, Hoffman-John, Lee, Aroyo, Rajakumar, Butryna, Lamm, Kuzmina, Fenton, Cohen, Bernstein, Kurzweil, Aguera-Arcas, Cui, Croak, Chi, and Le}]{Thoppilan2022LaMDALM}
Romal Thoppilan, Daniel~De Freitas, Jamie Hall, Noam~M. Shazeer, Apoorv Kulshreshtha, Heng-Tze Cheng, Alicia Jin, Taylor Bos, Leslie Baker, Yu~Du, Yaguang Li, Hongrae Lee, Huaixiu~Steven Zheng, Amin Ghafouri, Marcelo Menegali, Yanping Huang, Maxim Krikun, Dmitry Lepikhin, James Qin, Dehao Chen, Yuanzhong Xu, Zhifeng Chen, Adam Roberts, Maarten Bosma, Yanqi Zhou, Chung-Ching Chang, I.~A. Krivokon, Willard~James Rusch, Marc Pickett, Kathleen~S. Meier-Hellstern, Meredith~Ringel Morris, Tulsee Doshi, Renelito~Delos Santos, Toju Duke, Johnny~Hartz S{\o}raker, Ben Zevenbergen, Vinodkumar Prabhakaran, Mark D{\'i}az, Ben Hutchinson, Kristen Olson, Alejandra Molina, Erin Hoffman-John, Josh Lee, Lora Aroyo, Ravi Rajakumar, Alena Butryna, Matthew Lamm, V.~O. Kuzmina, Joseph Fenton, Aaron Cohen, Rachel Bernstein, Ray Kurzweil, Blaise Aguera-Arcas, Claire Cui, Marian~Rogers Croak, Ed~H. Chi, and Quoc Le. 2022.
\newblock \href {https://api.semanticscholar.org/CorpusID:246063428} {Lamda: Language models for dialog applications}.
\newblock \emph{ArXiv}, abs/2201.08239.

\bibitem[{Zhang* et~al.(2020)Zhang*, Kishore*, Wu*, Weinberger, and Artzi}]{bert-score}
Tianyi Zhang*, Varsha Kishore*, Felix Wu*, Kilian~Q. Weinberger, and Yoav Artzi. 2020.
\newblock \href {https://openreview.net/forum?id=SkeHuCVFDr} {Bertscore: Evaluating text generation with bert}.
\newblock In \emph{International Conference on Learning Representations}.

\end{thebibliography}

\appendix

\section{\ourmethod{} Methods Details} \label{sec:laq_methods_details}

In this section, we provide a full description of the \ourmethod{} methods used, described in \cref{sec:frameworks}.

\begin{table}[t]
    \centering
    \small
    \adjustbox{max width=\linewidth}{
        \begin{tabular}{llcc}
            \toprule
             & Method & Input Length & Output Length \\
             \midrule
             \parbox[t]{2mm}{\multirow{3}{*}{\rotatebox[origin=c]{90}{MDS}}} & \textsc{Vanilla} & 25674.7 \tiny{$\pm{396.7}$} & 227.7 \tiny{$\pm{6.6}$} \\
             & \textsc{ALCE} & 22239.6 \tiny{$\pm{279.5}$} & 214.5 \tiny{$\pm{6.8}$} \\
             & \textsc{Attr. First} & 2843.0 \tiny{$\pm{7.3}$} & 89.3 \tiny{$\pm{2.0}$} \\
             \toprule
             \parbox[t]{2mm}{\multirow{3}{*}{\rotatebox[origin=c]{90}{LFQA}}} & \textsc{Vanilla} & 58299.8 \tiny{$\pm{1031.1}$} & 232.4 \tiny{$\pm{8.0}$} \\
             & \textsc{ALCE} & 45104.4 \tiny{$\pm{826.6}$} & 200.9 \tiny{$\pm{8.1}$} \\
             & \textsc{Attr. First} & 3025.4 \tiny{$\pm{7.7}$} & 107.2 \tiny{$\pm{3.4}$}  \\             
             \bottomrule
        \end{tabular}
    }
    \caption{Average number of characters in the LLM prompt \ourmethod{} method, including standard error of the mean.}
    \label{tab:size_of_prompts}
\end{table}

\begin{figure}[t]
\begin{mylisting}{promptStyle}
<@\textcolor{instructionsColor}{You are provided with an output sentence and the source texts from which it was generated. You need to identify spans in the source from which the output sentence was generated. Copy verbatim the attributing source spans, and use a semicolon (;) as a delimiter between each consecutive span. The output sentence should be fully supported by the concatenation of the attributed source spans. IMPORTANT: Each span must be verbatim copied from the corresponding sources. Do not make any changes or paraphrases to the source spans. If necessary, you may copy multiple spans from the same or source, but avoid adding un-necessary spans and keep each span as short as possible.}@>

Input:
<@\color{black}Source 1: \underline{Voters in 11 states will pick their} \underline{governors tonight} Republicans appear on track to increase their numbers by at least one, and with the potential to extend their hold to more than two-thirds of the nation's top state offices@>
...

<@\color{black}Output: There is a race for the governor's mansion in 11 states today.@>

<@\color{red}Attribution: Voters in 11 states will pick their governors tonight@>
\end{mylisting}
\caption{Example prompt for LLM-based post-hoc alignment. The instructions are depicted in green, input to the model in black, and model's output in red. This example is one of three few-shot examples. The source texts of the few-shot examples are adapated based on the generation method: Vanilla includes all documents, ALCE includes relevant documents, and \textsc{Attr. First} includes relevant source spans.}
\label{fig:llm_based_alignment_prompt}
\end{figure}

\subsection{LLM Prompt}
The prompt is provided in \cref{fig:llm_based_alignment_prompt}. The average size of prompts is reported in \cref{tab:size_of_prompts}.
We use GPT-4o \citep{openai2024gpt4ocard}. In our experiments, we include three in-context examples sourced from the dev split of the corresponding datasets. We manually optimized the prompt instructions and few-shot examples based on iterations on the development set.

\subsection{LLM Internals}
Our LLM-based internals method is based on the method by \citet{phukan-etal-2024-peering}. Since this method requires access to the weights of the model, we run \textsc{Llama-3.1-8b-Instruct} on a single A100-80GB GPU for approximately 8 hours. More running time details are available in \cref{tab:llm_internals_time}.

\begin{table}[t]
    \centering
    \small
    \adjustbox{max width=\linewidth}{
        \begin{tabular}{llc}
            \toprule
             & Method & Avg. time (sec.) \\
             \midrule
             \parbox[t]{2mm}{\multirow{3}{*}{\rotatebox[origin=c]{90}{MDS}}} & \textsc{Vanilla} & 0.6 $\pm{0.0}$ \\
             & \textsc{ALCE} & 4.2 $\pm{0.2}$ \\
             & \textsc{Attr. First} & 8.6 $\pm{0.6}$ \\
             \toprule
             \parbox[t]{2mm}{\multirow{3}{*}{\rotatebox[origin=c]{90}{LFQA}}} & \textsc{Vanilla} & 0.7 $\pm{0.0}$ \\
             & \textsc{ALCE} & 20.1 $\pm{2.1}$ \\
             & \textsc{Attr. First} & 54.1 $\pm{4.0}$ \\
             \bottomrule
        \end{tabular}
    }
    \caption{Average time of the LLM internals \ourmethod{} method, including standard error of the mean.}
    \label{tab:llm_internals_time}
\end{table}

We now provide a short description of this work, and the adaptation we made to support the \ourmethod{} setting. The method proposed by \citet{phukan-etal-2024-peering} is based on the idea that LLMs have inherent awareness of the document parts they use while generating answers. They claim that it is likely captured by the hidden states of the LLM. Accordingly, their method includes creating a prompt that concatenates the query $q$, the documents $D$, and the output $o$, and then feeds this to a LLM in a single forward pass. This creates the hidden representations of the text.

Formally, the prompt is denoted $P$, such that  $P=q+D+o$, where `+' denotes concatenation. Also, the hidden layer representation of token $t_i\in{P}$ for layer $l$ of the model is denoted $h_i^l$. The attribution process is then composed of two sub-tasks:

\paragraph{Sub-Task 1: Identification of extractive answer tokens}
An important claim made in their paper is that not all tokens should be attributed, because some tokens are `glue' tokens created by the LLM. This task involves identifying extractive tokens, which are tokens that originate from the source documents, usually verbatim.

Formally, a token $o_i\in{o}$ is an extractive token if there exists a token $d_j\in{D}$ such that the cosine similarity between $h_i^l$ and $h_j^l$ is greater than a threshold $\theta$.

In our work, we use the threshold $\theta=0.7$ and layer $l=5$, which achieves the highest F1 scores based on their paper. In addition, as formalized in \cref{sec:IOLA_task_definition}, we only look at output spans $o_1,\ldots{},o_n$ provided as input, and not the entire output $o$.

\paragraph{Sub-Task 2: Attribution of extractive answer span S}

Given an output span $S$ with tokens $o_1,\ldots{},o_m\subseteq{o}$, compute the average hidden layer representation $h_S$ for each token $o_i\in{S}$ as:
$$
h_s=\frac{1}{n}\sum_{i=1}^{n}h_i^l
$$

Next, $h_s$ is used to identify anchor tokens in $D$. Anchor tokens, denoted $D_T$, are the tokens most similar to the output span $S$. This is calculated for each document token $d_j\in{D}$ as the cosine similarity between $h_S$ and $h_j^l$. For each anchor token $d_a\in{D_T}$, a window of tokens around $d_a$ is explored, up to a length $L$. For each window, an average representation is calculated and the highest ranked window is considered the attribution for $S$. In our work, we use $L=30$.

\section{Generation Methods Details} \label{sec:generation_methods_details}

In this section, we provide a full description of the generation methods used, described in \cref{sec:frameworks}.

As as a pre-processing step, we first decontextualize the output spans. We use the decontextualization prompt from MolecularFacts \citep{gunjal-durrett-2024-molecular}, which takes the concatenated output spans as input, together with the entire output as context, and outputs decontextualized facts. We used the original MolecularFacts prompt and ran it with GPT-4o. The resultant decontextualized fact is then mapped back to the output, as described in \cref{sec:alignment_factscore_to_spans}.

\subsection{ALCE}
\citet{gao-etal-2023-enabling} introduced the idea of allowing LLMs to generate citations together with the output. We use the same prompt as the original paper with two few-shot examples and $T=0.5$, following \citet{slobodkin-etal-2024-attribute}.

\subsection{Attr. First}
\citet{slobodkin-etal-2024-attribute} decompose the generation process into multiple explicit steps, allowing for precise attribution tracing. The first step, content selection, involves highlighting relevant source spans. The second step, sentence planning, consists of clustering spans for each sentence, followed by sentence generation based on the clustered information. Each new sentence is generated with conditioning on the previously generated sentences. We adopt the same prompt and few-shot demonstration examples as used in the original paper. Among the multiple variants of \textsc{Attr. First}, we select \textsc{Attr. First}$_{CoT}$, which the paper identifies as the best-performing variant.

\section{Experimental Setup Details}\label{sec:experimental_setup_details}

\subsection{Datasets}
Our benchmark includes both a multi-document summarization setting (MDS), as well as a long-form QA setting (LFQA). Both are content-grounded settings such that the source texts are used to generate an ouput. Specifically, we use SPARK \citep{ernst-etal-2024-power} for MDS, and the RAG-based dataset curated by \citet{liu-etal-2023-evaluating} for LFQA. We used the same split of the datasets created by \citet{slobodkin-etal-2024-attribute}. The datasets sizes are provided in \cref{tab:datasets_stats}. Both datasets are in English. The licenses for the datasets are following: \citet{ernst-etal-2024-power} CC BY-SA 4.0, \citet{liu-etal-2023-evaluating} MIT license.

\subsubsection{Synthesizing Attribution Queries}
Following \cref{sec:datasets}, we provide more details about the decomposition of an output text to output facts. We first split the output into sentences.\footnote{using spaCy \url{https://spacy.io/}} For each output sentence, we then run a prompt decomposing the output into atomic facts. FActScore \citep{min-etal-2023-factscore} is an LLM-based method used to breakdown a sentence into atomic facts. It is a prompt comprised of instructions and multiple few-shot examples. We used the original FActScore prompt and run it with GPT-4o. The resultant fact is then mapped back to the output, as described in \cref{sec:alignment_factscore_to_spans}.

\subsection{Evaluation}\label{sec:evaluation_details}
For calculating AutoAIS, we use the model \textsc{google/t5\_xxl\_true\_nli\_mixture} \citep{honovich-etal-2022-true-evaluating}, which is trained on NLI datasets and has been used in previous work to analyze attribution \citep{gao-etal-2023-rarr, slobodkin-etal-2024-attribute}. It correlates well with AIS scores  \citep{gao-etal-2023-rarr}.

\begin{table}[t]
    \centering
    \small
    \adjustbox{max width=\linewidth}{
        \begin{tabular}{llcc}
            \toprule
             Task & Dataset & Dev & Test \\
             \midrule
             MDS & \textsc{SPARK \citep{ernst-etal-2024-power}} & 45 & 65  \\
             \toprule
             LFQA & \textsc{Evaluating \citep{liu-etal-2023-evaluating}} & 44 & 45  \\
             \bottomrule
        \end{tabular}
    }
    \caption{Datasets sizes used in our benchmark for development and evaluation.}
    \label{tab:datasets_stats}
\end{table}

\section{Attribution Metadata Details}

Illustrated in \cref{fig:contributions}, we suggest that some generation methods can provide attribution metadata. In this section, we discuss the attribution metadata provided by the \textsc{Attr. First} method. Each sentence-level localized attribution is composed of one or more records, each consisting of the following information: output sentence idx, source file ID, and a list of source character offsets. For example, `<0, doc\_1.txt, [[17367, 17562]]>`. In comparison to non-localized attribution, such as the ALCE method, this requires one additional column for offsets. We analyze the average storage required for saving sentence-level attribution per output. Our analyses show that it requires 2Kb on average per attributed output, totalling in a fairly small increase of 700 bytes per attributed output.

\section{Alignment of Facts to Spans}
\label{sec:alignment_factscore_to_spans}

Throughout our work, we extracted facts from the output text and later needed to map them back to their corresponding spans. In this section, we describe the algorithm used to align extracted facts with the original output text.

The first application of this alignment process is in our evaluation methodology, where we decompose each output sentence into atomic facts using an LLM, as detailed in \cref{sec:datasets}. For instance, consider the sentence ``Exposing students to texts from different religions promotes understanding'' from \cref{tab:alignment_of_decomposed_facts}. To simulate a user's highlight, we need to align these atomic facts with spans in the output, providing the necessary spans for the \ourmethod{} method. In this example, the aligned highlight would be ``exposing students to texts from different religions \ldots{} promoting understanding.''

The second application is in our proposed method, where we decontextualize queries. For example, in \cref{tab:decontextualization_examples}, we need to align the fact ``The 911 calls were made around 4:30'' with the output text ``The \ldots{} 911 calls around 4:30 p.m.'' This alignment is crucial to ensure proper attribution, such as correctly highlighting the word ``911.''

To achieve this alignment, we implement a naive lexical alignment algorithm. This approach is expected to perform well since each output fact is extracted from a single output sentence, and the generated fact does not contain any paraphrases.

\eh{todo -- To assess its effectiveness, we evaluate the algorithm on 20 randomly selected examples, measuring its accuracy in correctly aligning extracted facts with their corresponding spans in the output text.}

Formally, given an output $o$ and a fact $f$ expressed by $o$, we wish to find spans $o_1,\ldots,o_n\subseteq{o}$ such that $f\models{concat(o_1,\ldots{},o_n})$. 
\paragraph{Alignment algorithm}
\begin{enumerate}
    \item{\textbf{Tokenization \& Lemmatization:} We first split the output $o$ into words $o_1,\ldots,o_n$, and the fact $f$ into words $f_1,\ldots,f_m$. Each word is lemmatized.\footnote{using spaCy \url{https://spacy.io/}}}
    \item{\textbf{Edit Script Calculation:}  We compute the edit script\footnote{Using Levenshtein distance \url{https://nltk.org/}} between the output words and the fact words. The edit script represents the minimal set of operations (insertions, deletions, and substitutions) required to transform one sequence into the other. Each word in the output is assigned an edit operation.}
    \item {\textbf{Word Alignment Based on Edit Operations:} Any output word $o_i$ classified as unchanged is considered aligned to the corresponding fact word $f_j$.}
\end{enumerate}

The advantage of using an edit script is that it considers the order in which the words appeared. However, sometimes the fact transposes information from the output sentence. For example, in the second row of \cref{tab:alignment_of_decomposed_facts}, the fact mentions ``public school'' after the mention of ``the First Amendment'', but in the output sentence the order is reversed. The algorithm will then not be able to align ``public school''. To support such transpositions, we generate a new fact $f'$ with non-aligned words from $f$. We then run this algorithm recursively with $f'$.

Overall, in 88\% of the examples we are able to align all content words,\footnote{Excluding stop-words \url{https://nltk.org}} and in 99\% we are able to align all content words but one.

\begin{table*}
    \small
    \centering
    \setlength{\tabcolsep}{0.5em}
    \renewcommand{\arraystretch}{1.4}
    \adjustbox{max width=\textwidth}{
        \begin{tabular}{p{2.0cm}p{12.4cm}}
            \toprule
             & Example \\
             \midrule

             Output sentence & The confirmation hearings for Brett Kavanaugh were marked by controversy over the withholding of documents, with \myyellow{Democrats} repeatedly \myyellow{complaining that Republicans} and the White House \myyellow{were keeping important records} from the public and the committee. \\
             \midrule
             LLM Prompt & Such theatrics have characterized Kavanaugh's hearings, in which \textbf{\myyellow{Democrats have repeatedly complained that Republicans have withheld documents from the committee and the public that shed important light on Kavanaugh's past}}. \ldots{} \myyellow{Democrats have repeatedly complained that the White House is withholding tens of thousands of documents relevant to the nomination} and wants many more that have been provided released to the public. \\
             LLM Internals & Such theatrics have characterized Kavanaugh's hearings, in which \textbf{Democrats} \myyellow{\textbf{have repeatedly complained that Republicans have withheld documents from the committee and the public that shed important light on Kavanaugh's past}}. \\
             \bottomrule
        \end{tabular}
    }
    \caption{Example MDS result. Top: one example output sentence from the \textsc{Attr. First} baseline with synthesized \ourmethod{} \myyellow{highlights}. Bottom: the predicted \myyellow{attributions}, with correct attribution in \textbf{bold}.} 
    \label{tab:failure_examples}
\end{table*}

\begin{table*}
    \centering
    \setlength{\tabcolsep}{0.5em}
    \renewcommand{\arraystretch}{1.2}
    \adjustbox{max width=\textwidth}{
        \begin{tabular}{p{3cm}p{11cm}}
            \toprule
             & Example \\
             \midrule
             Output sentence & \myyellow{The upcoming Supreme Court term is poised to address several contentious issues} that could significantly impact American society and politics. \\
             \midrule
             LLM prompt & After a year in which liberals scored impressive, high-profile \textbf{Supreme Court} victories, \myyellow{conservatives could be in line for wins on some of \textbf{this term's most contentious issues}}, as the justices consider cases that could gut public sector labor unions and roll back affirmative action at state universities.  \ldots{} \myyellow{A potential body blow to labor} Public-employee unions and politicians of both parties are keenly focused on a California dispute about whether states can compel government employees to pay union dues. \ldots{} Higher ed affirmative action back in the crosshairs \ldots{} \myyellow{The meaning of "one person, one vote'} \ldots{} Testing when abortion clinic regulations go too far \ldots{} \myyellow{The death penalty is shaping up to be a big issue for the Supreme Court as it begins a new term} \\
             LLM internals & However, \myyellow{as the \textbf{court's new term kicks off} Monday, uncertainty surrounds several other politically potent cases that could wind up on the court's agenda. Litigation over} state efforts to limit abortion by regulating clinics and doctors is making its way to the high court. Lois Lerner should have been gone shortly after the scandal first unraveled. \\
             \bottomrule
        \end{tabular}
    }
    \caption{Example MDS result. Top: one example output sentence from the Vanilla baseline with synthesized \ourmethod{} \myyellow{highlights}. Bottom: the predicted \myyellow{attribution}, with correct attribution in \textbf{bold}.}
    \label{tab:results_localization_examples_e2e}
\end{table*}

\end{document}